\documentclass[letterpaper,10pt,journal,twoside]{IEEEtran}

\IEEEoverridecommandlockouts                              % This command is only needed if 
\usepackage{amsmath}
\usepackage{amsmath,amsfonts}
\usepackage{algorithmicx}
\usepackage{algorithm}
\usepackage{array}
\usepackage[caption=false,font=normalsize,labelfont=sf,textfont=sf]{subfig}
\usepackage{textcomp}
\usepackage{stfloats}
\usepackage{url}
\usepackage{amsmath}
\usepackage{amssymb}
\usepackage{color}
\usepackage{algpseudocode}
\usepackage{xcolor}
\usepackage{makecell}
\usepackage{verbatim}
\usepackage{graphicx}
\usepackage{cite}
\usepackage{booktabs}
\usepackage{multirow}
\usepackage{amssymb}
\usepackage{fancyhdr}
\usepackage[table]{xcolor} 
\usepackage[colorlinks,
            linkcolor=blue,
            anchorcolor=blue,
            citecolor=blue]{hyperref}

\title{\LARGE \bf
SIPTraj: Map-Free End-to-End Trajectory Prediction via Physics-Guided Scene Interaction
}
\author{Feifei Liu,
Zejun Wei,
Haozhe Wang,
Yazhi Ye,
Yuying Zhang,
Jintao Cheng,
Chi Man Vong,
Xieyuanli Chen,~\IEEEmembership{Member,~IEEE,}
Xiaoyu Tang$^{*}$,~\IEEEmembership{Member,~IEEE}
\thanks{$^{*}$Corresponding author.}
\thanks{Feifei Liu, Zejun Wei, Haozhe Wang, Yazhi Ye and Yuying Zhang are with the School of Data Science and Engineering, Xingzhi College, South China Normal University, Shanwei, 516600, China. (e-mail: {\tt\small \{20238331056, 202481313613, 202481324245, 20258131088, 20258131189\}@m.scnu.edu.cn})}
\thanks{Jintao Cheng is with the Department of Electronic and Computer Engineering, Hong Kong University of Science and Technology, Hong Kong, China. (e-mail: {\tt\small jchengau@connect.ust.hk})}
\thanks{Chi Man Vong is with the University of Macau, Macau, China. (e-mail: {\tt\small cmvong@um.edu.mo})}
\thanks{Xieyuanli Chen is with the College of Intelligence Science and Technology, National University of Defense Technology, Changsha, China. (e-mail: {\tt\small xieyuanli.chen@nudt.edu.cn})}
\thanks{Xiaoyu Tang is with the School of Electronic Science and Engineering, South China Normal University, Foshan, Guangdong, 528225, China. (e-mail: {\tt\small tangxy@scnu.edu.cn})} }

\usepackage{amssymb}
\usepackage{amsmath}
\usepackage{tikz}
\usetikzlibrary{arrows.meta,positioning,shapes.geometric}
\begin{document}

\maketitle
\thispagestyle{empty}
\pagestyle{empty}

%%%%%%%%%%%%%%%%%%%%%%%%%%%%%%%%%%%%%%%%%%%%%%%%%%%%%%%%%%%%%%%%%%%%%%%%%%%%%%%%
\begin{abstract}
Trajectory prediction of surrounding agents is a prerequisite for safe 
planning and decision making in autonomous driving. Without high-definition 
(HD) maps, sensor-derived bird's-eye-view (BEV) features provide no explicit 
lane topology or drivable-area priors, making it inherently difficult to ground 
each agent in its surrounding scene context. Moreover, physical feasibility 
remains difficult to capture through data-driven learning alone, as kinematic 
constraints on agent motion cannot be explicitly encoded without structured 
supervision. Existing map-free predictors extract scene context in an 
agent-agnostic manner through a single fusion step and treat physical 
constraints only as output-level penalties, leaving both challenges unaddressed. 
We propose SIPTraj, a map-free trajectory prediction framework that jointly 
addresses scene grounding and physical feasibility. SIPTraj introduces a 
Hierarchical Agent-Scene Encoder (HASE) progressively grounding each agent 
in agent-guided scene evidence and refining inter-agent relations within the 
scene-grounded space. To tackle physical infeasibility in predicted trajectories, 
we develop a Physics-Guided Iterative Decoder (PGID). It conditions decoding 
on instantaneous kinematic states, propagating physical supervision into 
internal representations rather than output trajectories alone. Extensive 
experiments on nuScenes and Argoverse 2 Sensor show that SIPTraj surpasses 
prior map-free predictors and strong map-based baselines without any HD map 
at inference. Our code will be released as open-source.
\end{abstract}


%%%%%%%%%%%%%%%%%%%%%%%%%%%%%%%%%%%%%%%%%%%%%%%%%%%%%%%%%%%%%%%%%%%%%%%%%%%%%%%%
%%%%%%%%%%%%%%%%%%%%%%%%%%%%%%%%%%%%%%%%%%%%%%%%%%%%%%%%%%%%%%%%%%%%%%%%%%%%%%%%
\section{INTRODUCTION}

Accurate trajectory prediction of surrounding agents, which include 
vehicles, pedestrians, and cyclists, is a prerequisite for safe 
planning and decision making in autonomous driving 
\cite{caesar2020nuscenes,wilson2023argoverse,cheng2024mfmos}. The task is intrinsically challenging because future motion depends jointly on road structure, multi-agent interaction, and each agent's kinematic state. A reliable predictor must reason over these factors simultaneously and produce trajectories suitable for downstream planning.  
\begin{figure}[t]
  \centering
  \includegraphics[width=\columnwidth]{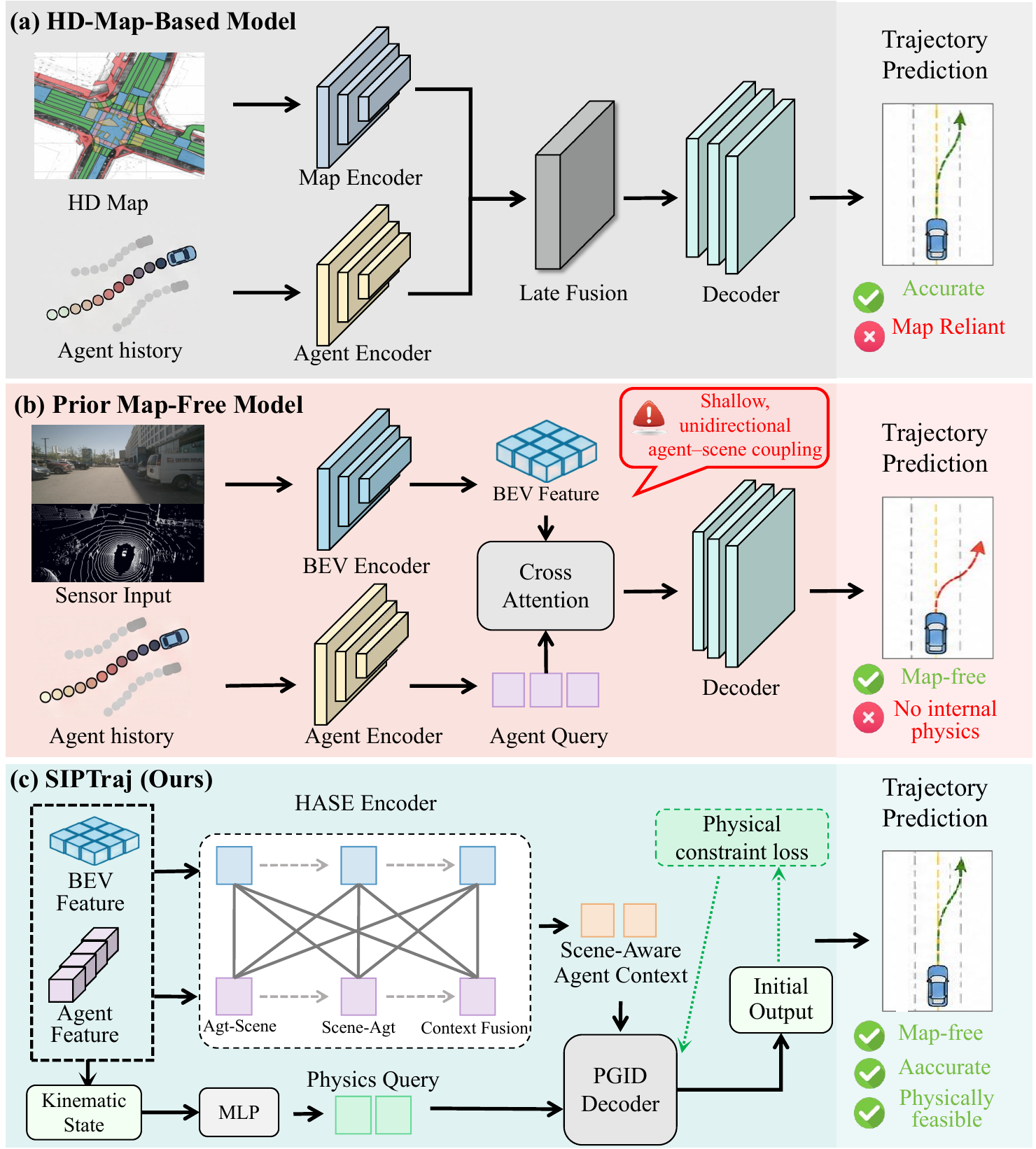}
  \caption{Comparison of trajectory prediction paradigms. 
    \textbf{(a)}~HD-map-based models achieve high accuracy but suffer from heavy HD-map dependency. 
    \textbf{(b)}~Prior map-free models replace maps with BEV features but rely on late fusion, leading to agent-agnostic scene extraction and physically unmodeled trajectories. 
    \textbf{(c)}~SIPTraj coordinates scene and agent streams via hierarchical self- and cross-interactions, and refines representations by injecting instantaneous kinematics as physics queries under consistency loss feedback, yielding map-free, accurate, and feasible predictions.}
  \label{fig:teaser}
  \vspace{-5mm}
\end{figure}

Existing trajectory prediction methods can be broadly categorized according to their use of high-definition (HD) maps. Map-based predictors exploit vectorized lane geometry and road topology as structured priors, achieving strong performance through graph or transformer architectures \cite{gao2020vectornet,liang2020lanegcn,shi2022mtr,zhou2023qcnet}. The reliance on accurate and up-to-date HD maps, however, limits scalability in real-world deployment where mapping quality and coverage cannot always be guaranteed. To reduce this dependency, most recent map-free trajectory prediction methods turn to sensor-derived bird's-eye-view (BEV) features extracted from cameras and LiDAR \cite{li2024bevformer,liu2023bevfusion} as a natural substitute for vectorized maps, predicting trajectories from end-to-end visual queries \cite{hu2023vip3d}, rasterized BEV inputs \cite{yadav2024caspformer}, or fused multi-modal BEV features \cite{kong2025bevtraj}. These map-free predictors, however, inherit the design pattern of map-based methods largely unchanged, overlooking two structural priors that HD maps once supplied implicitly but that BEV features cannot provide in the same form.

Prior map-free predictors demonstrate that sensor-derived BEV 
representations can serve as effective substitutes for HD maps. 
ViP3D \cite{hu2023vip3d} propagates per-agent visual queries 
end-to-end across time to enable fully differentiable trajectory 
prediction from raw camera inputs. CASPFormer 
\cite{yadav2024caspformer} recurrently decodes trajectories from 
rasterized multi-scale BEV scene encodings via deformable attention, 
and BEVTraj \cite{kong2025bevtraj} compresses dense BEV features into 
compact aggregated representations before fusing with agent history 
features. However, across these methods, the scene context available 
to each agent is extracted without conditioning on that agent's 
individual motion, and the interaction between scene representations 
and agent features remains a single-step fusion rather than iterative 
mutual refinement. Since BEV features encode spatial structure only 
implicitly, this shallow coupling leaves each agent's representation 
insufficiently grounded in its local spatial environment.

As noted above, prior map-free predictors also neglect physical 
feasibility as an explicit representational objective. Existing work 
either imposes no physical constraints or applies them only as 
output-level penalties on predicted trajectories 
\cite{gao2024sif,zhong2025litransformer,salzmann2020trajectron}, 
treating physics as a training signal rather than a learned prior. 
An agent's instantaneous kinematic state, however, directly constrains 
how its position can evolve regardless of scene layout, and encoding 
it as a learnable signal inside the model would allow physical 
regularities to shape internal representations rather than merely 
correct output trajectories.

To address these limitations, as illustrated in 
Fig.~\ref{fig:teaser}, we propose SIPTraj, a map-free trajectory 
prediction framework that recovers the structural priors lost when 
HD maps are removed. Our key insight is that BEV features, though 
lacking explicit spatial structure, contain rich perceptual 
information that can be progressively transferred into agent 
representations if scene context extraction is conditioned on each 
agent's motion and the resulting scene evidence is iteratively 
written back to agent features across multiple stages. Guided by 
this insight, we design a Hierarchical Agent-Scene Encoder (HASE), 
which conditions scene aggregation on agent motion context, grounds 
each agent's representation in its local BEV environment, and 
refines inter-agent interactions within the scene-grounded space, 
replacing the single-step fusion of prior map-free predictors. 
Beyond agent-scene coupling, the instantaneous kinematic state of 
each agent, including its speed, acceleration, and curvature, is a 
structured prior that constrains future trajectory evolution 
regardless of scene layout, yet prior work leaves it outside the 
model's internal representations. To exploit this, we further 
propose a Physics-Guided Iterative Decoder (PGID), which encodes 
each agent's kinematic state as a learnable query inside every 
decoder layer, allowing physical regularities to shape the internal 
scene-context representation rather than acting only on the output 
trajectory. Without relying on HD maps, SIPTraj matches or surpasses 
many state-of-the-art map-based predictors on the nuScenes and 
Argoverse~2 Sensor benchmarks \cite{caesar2020nuscenes,wilson2023argoverse}.

In summary, our contributions are threefold.

(i) We propose SIPTraj, a map-free trajectory prediction framework that explicitly compensates for the structural priors lost when HD maps are removed, by coupling scene-grounded encoding with kinematic-state-conditioned decoding under a unified design principle.

(ii) We design a hierarchical agent-scene encoder that grounds agent representations in BEV context across multiple stages and reasons inter-agent interaction within the scene-grounded space, replacing single-shot agent-scene fusion with progressive bidirectional coupling.

(iii) We introduce Physics-Aware Feature Conditioning, which conditions scene-context features on each agent's instantaneous kinematic state before decoding. This couples physical supervision with the internal representation rather than confining it to the output trajectory, and is complemented by acceleration, jerk, and curvature losses that further enforce trajectory feasibility during training.

\section{Related Work}

\subsection{Learning-based Trajectory Prediction}

Trajectory prediction has evolved from recurrent encoders such as Social LSTM \cite{alahi2016social} to graph-based representations and transformer-based architectures that jointly model multi-agent interaction and scene context. According to whether high-definition (HD) maps are used, recent methods can be broadly categorized into map-based and map-free approaches.

Map-based methods exploit vectorized lane geometry and road topology as structured priors. Early vectorized approaches such as VectorNet \cite{gao2020vectornet} and LaneGCN \cite{liang2020lanegcn} model agent-lane and lane-lane interactions through graph representations of map elements. Subsequent transformer-based predictors such as MTR \cite{shi2022mtr}, QCNet \cite{zhou2023qcnet}, HiVT \cite{zhou2022hivt}, AutoBot \cite{girgis2022autobot}, and Wayformer ~\cite{nayakanti2023wayformer} further improve multimodal forecasting by combining map encoding with agent-level attention. While these methods achieve strong benchmark performance, their dependence on accurate and up-to-date HD maps limits scalability in real-world deployment where map quality and coverage cannot always be guaranteed.

To reduce this dependency, most recent map-free trajectory prediction methods replace vectorized maps with sensor-derived bird's-eye-view (BEV) representations extracted from cameras and LiDAR. ViP3D \cite{hu2023vip3d} performs end-to-end visual trajectory prediction directly from multi-view video. CASPFormer \cite{yadav2024caspformer} predicts multimodal trajectories from rasterized BEV inputs using deformable attention. BEVTraj \cite{kong2025bevtraj} builds upon a multi-modal BEV backbone \cite{liu2023bevfusion} and aggregates BEV features through agent-conditioned deformable queries. SIPTraj follows this direction and further investigates how BEV context can be injected into agent representations across multiple stages of encoding rather than only at a single fusion module.

\subsection{Physical Priors and Constrained Trajectory Prediction}

Several lines of work seek to improve the physical plausibility of predicted trajectories. One direction adds output-level regularization, where smoothness, acceleration, or curvature penalties are imposed on the predicted trajectories during training \cite{gao2024sif,zhong2025litransformer}. A second direction integrates motion dynamics directly into the trajectory decoder. Trajectron++ \cite{salzmann2020trajectron} replaces the position regressor with a dynamically-feasible integrator that produces trajectories from predicted control inputs, ensuring kinematic consistency at the decoder output.

Both directions act at the output stage of the prediction pipeline, and the upstream scene-context representation is not explicitly conditioned on each agent's instantaneous kinematic state. SIPTraj is complementary to these efforts. In addition to applying acceleration, jerk, and curvature penalties to the predicted trajectory, SIPTraj conditions the scene-context representation on each agent's instantaneous kinematic state before decoding, so that physical supervision shapes the internal representation rather than acting only at the output level.
\section{Method}
%========================================================================
%========================================================================
% SIPTraj --- Method section (drop-in replacement for Section 3)
%========================================================================

\label{sec:method}
\begin{figure*}[t]
  \centering
  \includegraphics[width=\linewidth]{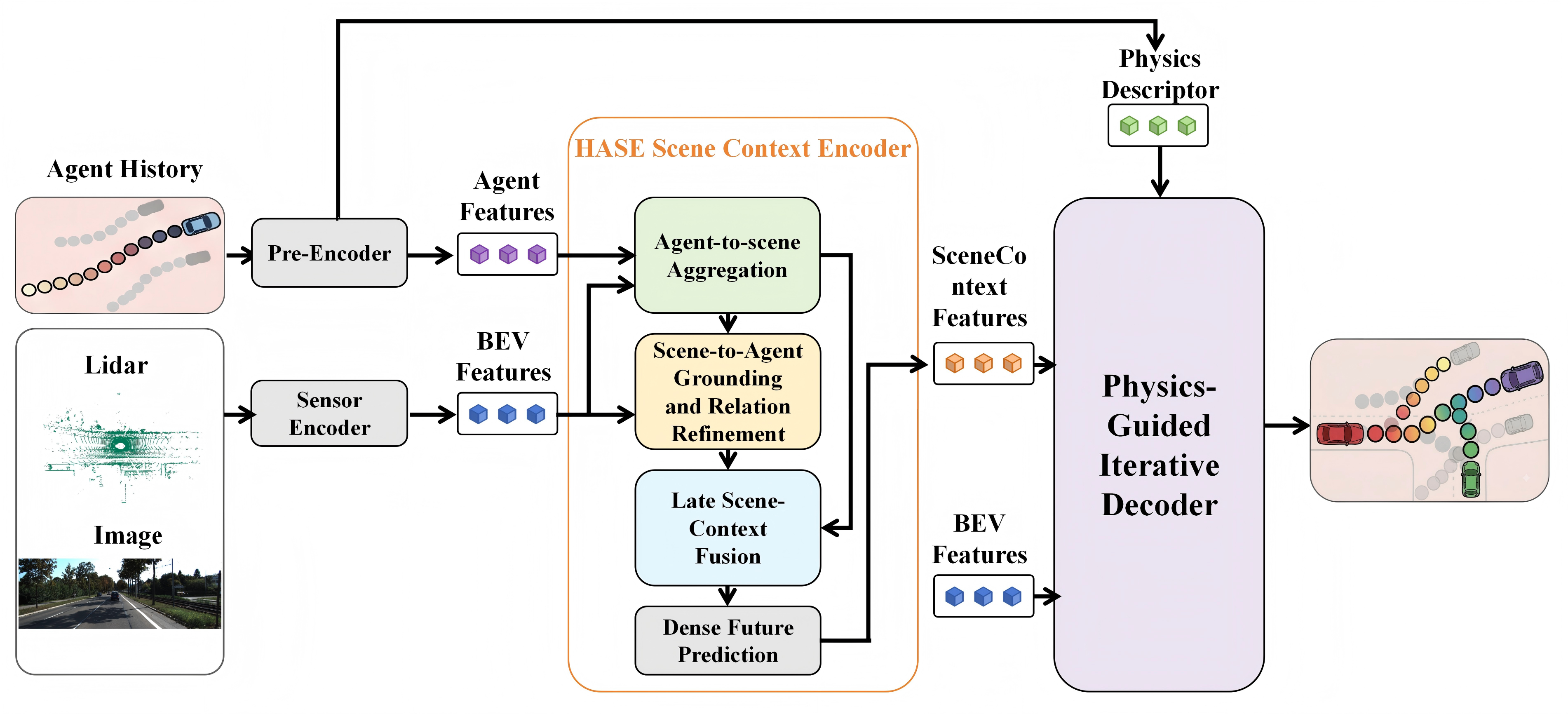}
  \caption{Overall architecture of SIPTraj. Sensor Encoder fuses multimodal
  sensor data (\textit{e.g.}, camera images, LiDAR point clouds) into BEV
  feature, while Pre-Encoder captures agent motion history. HASE Scene Context
  Encoder progressively couples the two streams through agent-to-scene
  aggregation, scene-to-agent grounding and relation refinement, and late
  scene-context fusion, producing scene context feature. Physics-Guided
  Iterative Decoder conditions scene context feature on a physics descriptor
  extracted from each agent's instantaneous kinematic state, and iteratively
  refines the target agent's multimodal trajectories using BEV feature.}
  \label{fig:overview}
\end{figure*}
In this section, we present SIPTraj, a map-free trajectory prediction framework that compensates for the structural priors lost when HD maps are absent through two complementary mechanisms: a Hierarchical Agent-Scene Encoder (HASE) that injects perceptual evidence into agent representations across multiple stages, and a Physics-Guided Iterative Decoder (PGID) that couples the scene 
representation to the agent's instantaneous kinematic state in a 
closed loop. An overview of SIPTraj is depicted in Fig.~\ref{fig:overview}.\subsection{Problem Formulation}
\label{sec:problem}

Let $\mathbf{X}\in\mathbb{R}^{N\times T_h\times C_x}$ denote the observed states of $N$ agents, including the prediction target, and $\mathbf{B}\in\mathbb{R}^{C_b\times H\times W}$ the BEV feature map from the sensor encoder. The agent pre-encoder maps $\mathbf{X}$ to agent tokens $\mathbf{F}=\{\mathbf{f}_i\}_{i=1}^{N}$, where $\mathbf{f}_i\in\mathbb{R}^D$. Given $(\mathbf{X},\mathbf{B})$, SIPTraj predicts $K$ multimodal future trajectory hypotheses for the target agent over $T_f$ future timesteps together with their categorical probabilities.

\subsection{Overall Framework}
\label{sec:overall}
Given multi-modal sensor inputs and the observed histories of surrounding agents, SIPTraj employs the following pipeline:
\textit{(1) Sensor Encoder.} A BEVFusion~\cite{liu2023bevfusion} backbone fuses camera and LiDAR inputs into a unified BEV feature map $\mathbf{B}\in\mathbb{R}^{C_b\times H\times W}$, which is shared by all downstream modules through deformable attention~\cite{zhu2021deformable}.
\textit{(2) Hierarchical Agent-Scene Encoder.} HASE progressively couples the motion histories $\mathbf{X}$ with the BEV representation $\mathbf{B}$. It first constructs the agent-guided scene memory $\mathbf{Z}(\mathbf{P}^{s})$, then grounds the pre-encoded agent tokens $\mathbf{F}$ in local scene context and refines their relations. Late Scene-Context Fusion produces the agent-indexed context $\mathbf{C}$ for downstream decoding.
\textit{(3) Physics-Guided Iterative Decoder.} PGID extracts a compact physical descriptor from each agent and uses it inside every decoder layer to produce an explicit coordinate correction on the predicted trajectory, while a parallel attention branch carries the physical signal into the mode content representation. Acceleration, jerk, and curvature consistency losses on the final trajectory backpropagate through both channels, shaping the internal physics encoding rather than only the output.

\begin{figure}[t]
\centering
\includegraphics[width=\linewidth]{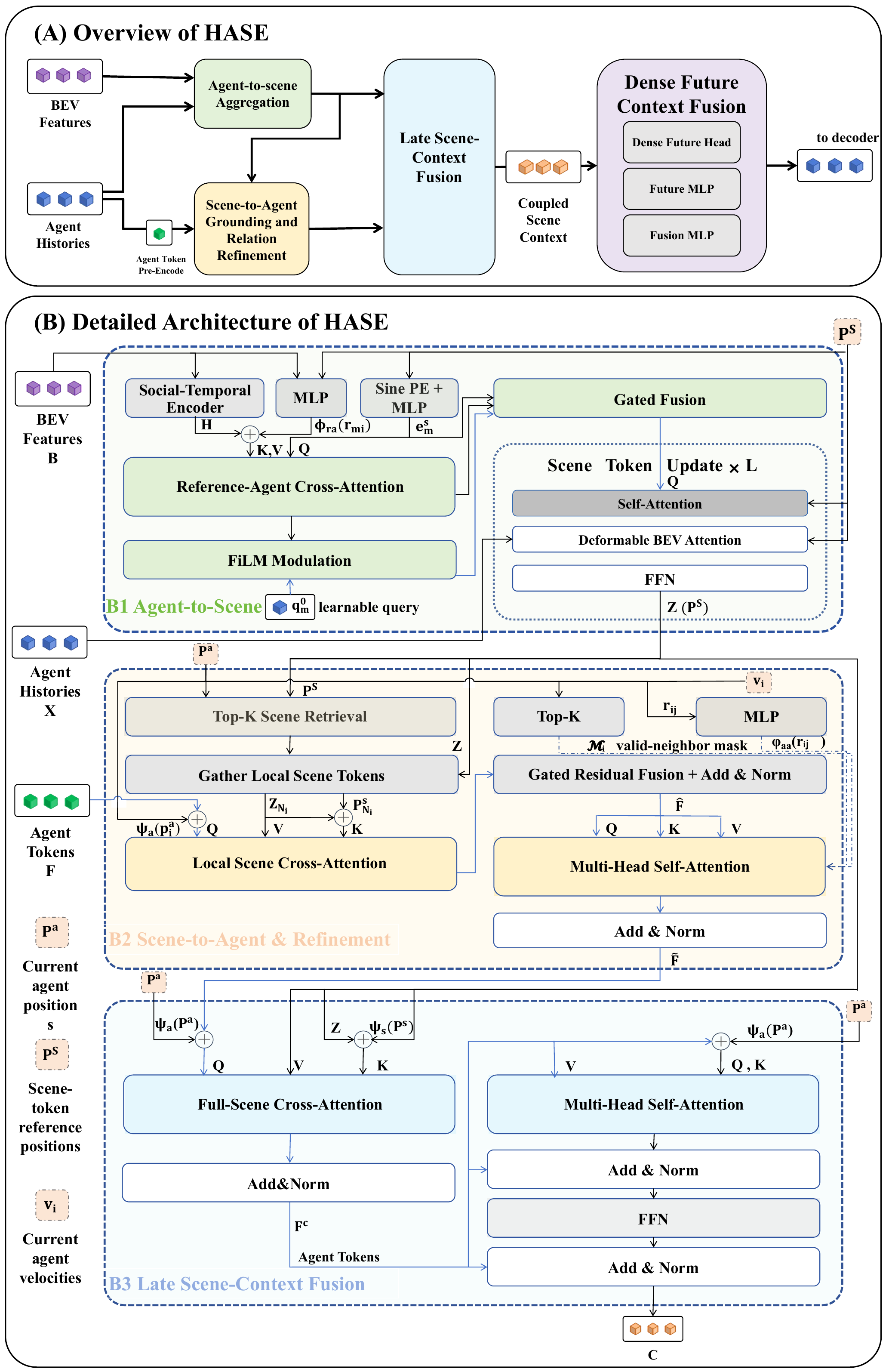}
\caption{Architecture of the Hierarchical Agent-Scene Encoder (HASE).
(A) Overview of the three coupling stages. (B) Detailed architecture.
Agent-conditioned queries read the BEV feature through deformable attention
to form agent-guided scene tokens (B1). Each agent token is then grounded in
its local scene memory and refined by relation-biased self-attention (B2),
and finally attends to the full scene memory to produce the coupled scene
context $\mathbf{C}$ (B3).}
\label{fig:hase}
\end{figure}

\subsection{Hierarchical Agent-Scene Encoder}
\label{sec:hase}

As illustrated in Fig.~\ref{fig:hase}, HASE maintains two parallel representations of the observed histories and couples them with the BEV representation through three stages: Agent-to-Scene Aggregation, Scene-to-Agent Grounding and Relation Refinement, and Late Scene-Context Fusion.

\subsubsection{Agent-to-Scene Aggregation}

We first extract agent-guided scene tokens from the dense BEV feature map. The observed histories are encoded into social-temporal agent context:
\begin{equation}
\mathbf{H}
=
\mathrm{Enc}_{\mathrm{st}}(\mathbf{X}),
\qquad
\mathbf{P}^{s}
=
\{\mathbf{p}^{s}_{m}\}_{m=1}^{M},
\qquad
\mathbf{Q}^{0}
=
\{\mathbf{q}^{0}_{m}\}_{m=1}^{M},
\end{equation}
where $\mathbf{P}^{s}$ and $\mathbf{Q}^{0}$ denote the scene-reference positions and learnable base queries, respectively.

Given a scene reference $\mathbf{p}^{s}_{m}$, we first obtain its reference embedding $\mathbf{e}^{s}_{m}=\phi_p(\mathrm{PE}(\mathbf{p}^{s}_{m}))$, where $\mathrm{PE}(\cdot)$ denotes sinusoidal positional encoding. To condition scene queries on agent motion, $\mathbf{e}^{s}_{m}$ queries the relation-conditioned agent context:
\begin{align}
\mathbf{c}_{m}
&= \operatorname{CrossAttn}\!\Bigl(
     \mathbf{e}^{s}_{m},\;
     \bigl\{\mathbf{h}_{i}+\phi_{\mathrm{ra}}(\mathbf{r}_{mi})\bigr\}_{i=1}^{N},
     \nonumber\\
&\qquad\qquad\qquad\quad\;
     \bigl\{\mathbf{h}_{i}+\phi_{\mathrm{ra}}(\mathbf{r}_{mi})\bigr\}_{i=1}^{N}
   \Bigr),
\\[4pt]
\mathbf{q}_{m}
&= \operatorname{GateFuse}\!\left(
     \operatorname{FiLM}\!\left(
       \mathbf{q}^{0}_{m},\,\mathbf{c}_{m}
     \right),\,
     \mathbf{c}_{m},\,
     \mathbf{e}^{s}_{m}
   \right).
\end{align}
Here, the three inputs of $\operatorname{CrossAttn}(\cdot)$ correspond to query, key, and value, respectively. The reference-agent descriptor $\mathbf{r}_{mi}$ is constructed from the scene reference and agent pose metadata, and $\phi_{\mathrm{ra}}(\cdot)$ maps it to a relation embedding. The resulting query $\mathbf{q}_{m}$ conditions the subsequent scene-token updates on agent motion.

The conditioned queries are updated through scene-token self-attention, deformable BEV attention, and a residual FFN:
\begin{equation}
\begin{aligned}
\mathbf{Q}^{(0)}
&=
\{\mathbf{q}_{m}\}_{m=1}^{M}, \quad \mathbf{Z}
=
\mathbf{Q}^{(L)}.\\
\mathbf{Q}^{(l+1)}
&=
\mathrm{SceneUpdate}_{l}
\left(
\mathbf{Q}^{(l)},
\mathbf{B},
\mathbf{P}^{s}
\right),\\
\end{aligned}
\end{equation}
The final output $\mathbf{Z}(\mathbf{P}^{s})$, where $\mathbf{Z}=\{\mathbf{z}_{m}\}_{m=1}^{M}\in\mathbb{R}^{M\times D}$, forms the agent-guided scene memory.

\subsubsection{Scene-to-Agent Grounding and Relation Refinement}

Given the pre-encoded agent tokens $\mathbf{F}$ and scene memory $\mathbf{Z}(\mathbf{P}^{s})$, each agent $i$ retrieves the $K_s$ scene tokens whose references are closest to its current position $\mathbf{p}^{a}_{i}$. The agent token with positional encoding is used as the query, while the retrieved scene tokens provide keys and values:
\begin{equation}
\mathbf{s}_{i}
=
\operatorname{Attn}
\left(
W_q(\mathbf{f}_{i}+\psi_a(\mathbf{p}^{a}_{i})),
W_k(\mathbf{Z}_{\mathcal{N}_{i}}+\psi_s(\mathbf{P}^{s}_{\mathcal{N}_{i}})),
W_v\mathbf{Z}_{\mathcal{N}_{i}}
\right).
\end{equation}
The local scene context is then written back through gated residual fusion:
\begin{equation}
\hat{\mathbf{f}}_{i}
=
\mathrm{LN}
\left(
\mathbf{f}_{i}
+
\sigma(W_g[\mathbf{f}_{i};\mathbf{s}_{i}])
\odot
\mathbf{s}_{i}
\right).
\end{equation}
For relation refinement, we retrieve the $K_n$ nearest candidate agents for each agent $i$ and mask invalid keys, obtaining a local valid-agent set $\mathcal{M}_{i}$. The descriptor $\mathbf{r}_{ij}$ encodes pairwise geometric and motion cues derived from the current agent positions $\mathbf{P}^{a}$ and velocities $\{\mathbf{v}_{i}\}$. For attention head $h$, the relation bias and spatial decay are injected into the attention score:
\begin{equation}
e_{ij}^{(h)}
=
\frac{
(W_Q^{(h)}\hat{\mathbf{f}}_{i})^{\top}
(W_K^{(h)}\hat{\mathbf{f}}_{j})
}{
\sqrt{D_h}
}
+
\phi_{\mathrm{aa}}^{(h)}(\mathbf{r}_{ij})
-
\frac{d_{ij}}{\sigma_d},
\qquad
j\in\mathcal{M}_{i}.
\end{equation}
The attended interaction update is scaled by $\alpha$ and added residually to $\hat{\mathbf{F}}$, producing the relation-refined representation $\tilde{\mathbf{F}}$.

\subsubsection{Late Scene-Context Fusion}

Scene-to-Agent Grounding exposes each agent to only a local subset of scene tokens. Late Scene-Context Fusion therefore allows the relation-refined agent tokens $\tilde{\mathbf{F}}$ to re-access the complete scene memory $\mathbf{Z}(\mathbf{P}^{s})$ before decoding.

Full-scene cross-attention first produces scene-aware agent tokens:
\begin{equation}
\mathbf{F}^{c}
=
\mathrm{LN}
\left(
\tilde{\mathbf{F}}
+
\operatorname{Attn}
\left(
\tilde{\mathbf{F}}+\psi_a(\mathbf{P}^{a}),
\mathbf{Z}+\psi_s(\mathbf{P}^{s}),
\mathbf{Z}
\right)
\right).
\end{equation}

Agent self-attention then uses $\mathbf{F}^{c}+\psi_a(\mathbf{P}^{a})$ as queries and keys and $\mathbf{F}^{c}$ as values. A residual FFN produces the coupled agent-context tokens $\mathbf{C}\in\mathbb{R}^{N\times D}$. Before decoding, the retained dense-future module predicts a coarse future for each valid agent and uses its encoded future feature to refine $\mathbf{C}$. For notational simplicity, the refined context passed to PGID is also denoted by $\mathbf{C}$.

\subsection{Physics-Guided Iterative Decoder}
\label{sec:pgid}

Map-free decoders rely on learned scene features to generate multimodal trajectories, but their mode queries and coordinate updates are not explicitly aware of the target agent's kinematic state. This may produce predictions that are visually plausible from the BEV representation but inconsistent with the agent's current speed, acceleration, or turning tendency. To address this issue, we propose the Physics-Guided Iterative Decoder (PGID), which conditions both mode initialization and iterative trajectory refinement on a compact physical descriptor. As shown in Fig.~\ref{fig:pgid}, PGID consists of three modules: Mode Initialization, Physics-Conditioned Attention Refinement (PCAR), and Iterative Physics-Guided Decoding. We denote the agent-indexed context supplied by HASE as $\mathbf{C}=\{\mathbf{c}_{i}\}_{i=1}^{N}\in\mathbb{R}^{N\times D}$, where each token is associated with one observed agent.

\subsection{Physics-Guided Iterative Decoder}
\label{sec:pgid}

\begin{figure}[t]
    \centering
    \includegraphics[width=\linewidth]{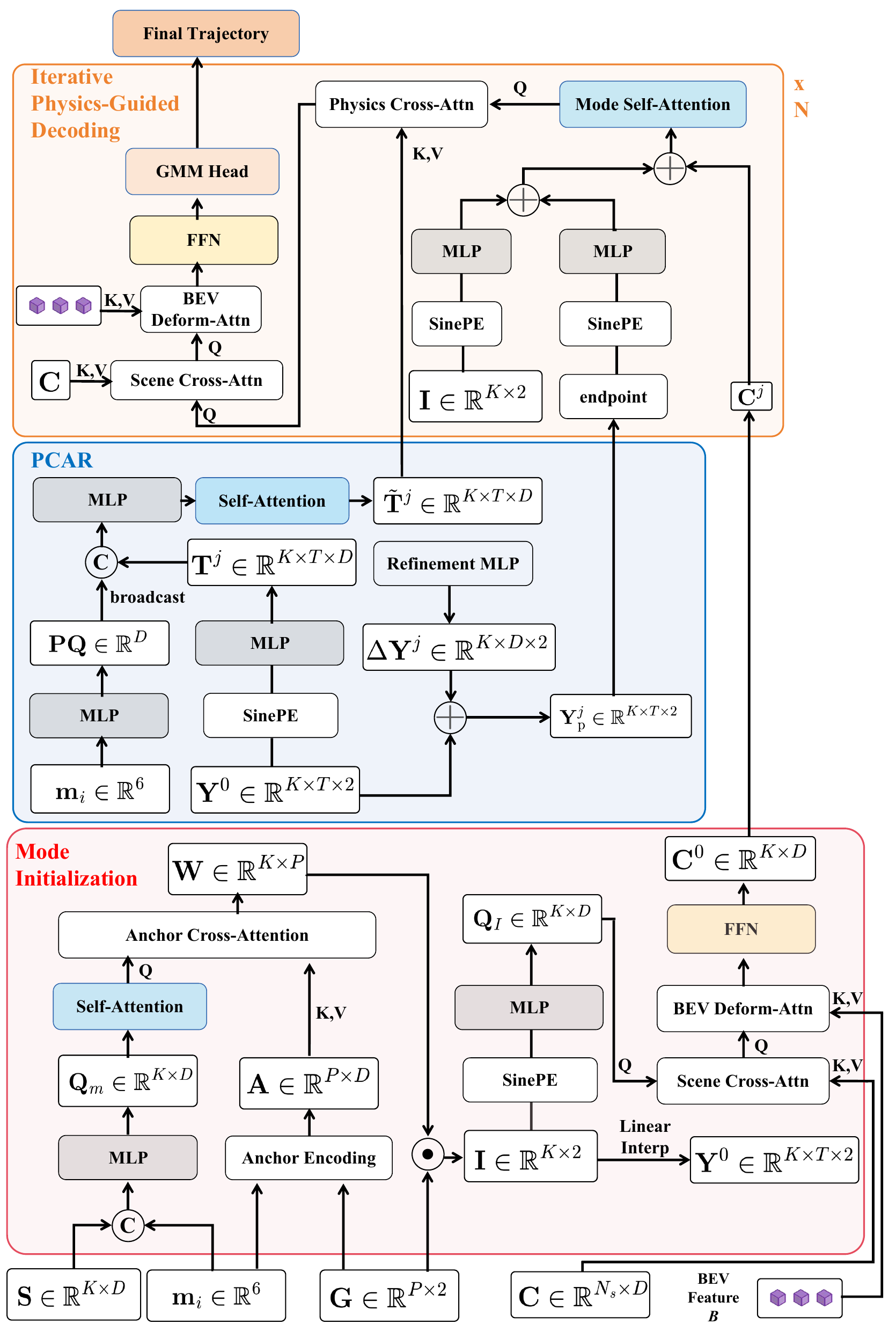}
    \caption{Architecture of the Physics-Guided Iterative Decoder (PGID). Mode Initialization generates dynamic intention points $\mathbf{I}$, initial content $\mathbf{C}^{0}$, and initial trajectories $\mathbf{Y}^{0}$ from the physical descriptor, scene context, and BEV features. PCAR uses the descriptor to produce a coordinate correction and physics-aware trajectory tokens. Iterative Physics-Guided Decoding then updates the mode content through mode self-attention, physics cross-attention, scene cross-attention, BEV deformable attention, and a GMM head. Physics consistency losses supervise the final trajectory and guide the internal physics-conditioned representations.}
    \label{fig:pgid}
\end{figure}

Map-free decoders rely on learned scene features to generate multimodal trajectories, but their mode queries and coordinate updates are not explicitly aware of the target agent's kinematic state. This may produce predictions that are visually plausible from the BEV representation but inconsistent with the agent's current speed, acceleration, or turning tendency. To address this issue, we propose the Physics-Guided Iterative Decoder (PGID), which conditions both mode initialization and iterative trajectory refinement on a compact physical descriptor. As shown in Fig.~\ref{fig:pgid}, PGID consists of three modules: Mode Initialization, Physics-Conditioned Attention Refinement (PCAR), and Iterative Physics-Guided Decoding. We denote the HASE output as $\mathbf{C}\in\mathbb{R}^{N_s\times D}$ in the decoder, where $N_s$ is the number of scene context tokens.

For target agent $i$, PGID summarizes the current physical state from the last valid observation. We use velocity $\mathbf{v}_i$, acceleration $\mathbf{a}_i$, heading $\hat{\mathbf{v}}_i$, and a curvature proxy derived from the lateral acceleration. The descriptor is written as
\begin{equation}
\mathbf{m}_i =
\big[
\bar{v}_i,
\bar{a}_{x,i},
\bar{a}_{y,i},
\hat{v}_{x,i},
\hat{v}_{y,i},
\bar{\kappa}_i
\big]^{\top}
\in\mathbb{R}^{6},
\label{eq:phys_descriptor}
\end{equation}
where $\bar{(\cdot)}$ denotes normalized values. The curvature term is computed from the component of acceleration orthogonal to the heading direction, so it exposes turning tendency while keeping straight motion close to zero.

\subsubsection{Mode Initialization}

Mode Initialization generates dynamic intention points and initial mode features before iterative decoding. It starts from learnable mode slots $\mathbf{S}\in\mathbb{R}^{K\times D}$ and candidate goal anchors $\mathbf{G}\in\mathbb{R}^{P\times 2}$. The physical descriptor $\mathbf{m}_i$ is embedded and fused with $\mathbf{S}$ through an MLP, producing physics-conditioned mode queries $\mathbf{Q}_m$. In parallel, $\mathbf{G}$ is encoded with positional information and its relation to $\mathbf{m}_i$, producing an anchor memory $\mathbf{A}$. The mode queries attend to $\mathbf{A}$ and generate anchor weights $\mathbf{W}$, which aggregate the goal anchors into dynamic intention points:
\begin{equation}
\mathbf{I} = \mathbf{W}\mathbf{G},
\qquad
\mathbf{I}\in\mathbb{R}^{K\times 2}.
\end{equation}

The dynamic intention points guide both trajectory initialization and feature initialization. PGID linearly interpolates from the target origin to $\mathbf{I}$ to obtain the initial trajectories $\mathbf{Y}^{0}\in\mathbb{R}^{K\times T\times 2}$. It also injects the positional encoding of $\mathbf{I}$ into $\mathbf{Q}_m$, then lets the resulting queries attend to the scene context $\mathbf{C}$ and the BEV feature map $\mathbf{B}$ through cross-attention and deformable attention. The resulting content feature $\mathbf{C}^{0}\in\mathbb{R}^{K\times D}$, together with $\mathbf{Y}^{0}$ and $\mathbf{I}$, initializes the following decoder layers.

\subsubsection{Physics-Conditioned Attention Refinement}

Following prior work\cite{shi2022motion}, PCAR converts the physical descriptor into a trajectory-level correction at each decoder layer. Given the current trajectory $\mathbf{Y}^{j}\in\mathbb{R}^{K\times T\times 2}$, it first maps $\mathbf{m}_i$ to a physics query $\mathbf{PQ}\in\mathbb{R}^{D}$ and tokenizes $\mathbf{Y}^{j}$ with sinusoidal positional encoding and an MLP. The trajectory tokens are coupled with $\mathbf{PQ}$ through concatenation and an MLP, followed by temporal self-attention. This produces physics-aware trajectory tokens $\tilde{\mathbf{T}}^{j}\in\mathbb{R}^{K\times T\times D}$ that carry both temporal geometry and physical state information.

A refinement MLP regresses a coordinate residual $\Delta\mathbf{Y}^{j}$ from $\tilde{\mathbf{T}}^{j}$. The trajectory is then updated by a residual connection:
\begin{equation}
\mathbf{Y}^{j}_{\mathrm{p}}
=
\mathbf{Y}^{j}
+
\Delta\mathbf{Y}^{j}.
\label{eq:pcar_residual}
\end{equation}
PCAR passes the corrected trajectory $\mathbf{Y}^{j}_{\mathrm{P}}$ and the tokens $\tilde{\mathbf{T}}^{j}$ to the next stage, allowing physics to influence both coordinates and mode representations.

\subsubsection{Iterative Physics-Guided Decoding}

Iterative Physics-Guided Decoding updates the mode content by combining intention, physics, scene context, and BEV evidence. It first builds a positional query from the dynamic intention point $\mathbf{I}$ and the endpoint of $\mathbf{Y}^{j}_{\mathrm{P}}$, and adds this query to the current mode content $\mathbf{C}^{j}$. The resulting feature is processed by mode self-attention, followed by physics cross-attention where $\tilde{\mathbf{T}}^{j}$ serves as keys and values. This branch lets each mode directly consult the physics-aware trajectory tokens produced by PCAR.

The updated mode content then attends to $\mathbf{C}$ to absorb scene-level interaction information. It further applies deformable attention over $\mathbf{B}$, using the endpoint of $\mathbf{Y}^{j}_{\mathrm{P}}$ as the BEV reference point. An FFN and a Gaussian Mixture Model(GMM) head finally produce the next-layer content $\mathbf{C}^{j+1}$, trajectory $\mathbf{Y}^{j+1}$, and mode probability $\boldsymbol{\pi}^{j+1}$. This process is repeated for $N$ decoder layers.

We further apply acceleration, jerk, and curvature consistency losses to the selected predicted trajectory. These losses are output-level penalties, but their gradients pass through the coordinate residual path and the physics cross-attention path, encouraging the internal representations to remain physically feasible. The total objective is
\begin{equation}
\mathcal{L}
=
\mathcal{L}_{\mathrm{task}}
+
\lambda_a\mathcal{L}_a
+
\lambda_j\mathcal{L}_j
+
\lambda_c\mathcal{L}_c,
\end{equation}
where $\mathcal{L}_{\mathrm{task}}$ is the multimodal prediction loss and $\lambda_a$, $\lambda_j$, and $\lambda_c$ are scalar weights.

\section{Experiments}
\label{sec:experiments}

\subsection{Datasets and Evaluation Metrics}

\subsubsection{Datasets}
We evaluate SIPTraj on two large-scale autonomous driving benchmarks: nuScenes~\cite{caesar2020nuscenes} and Argoverse~2 Sensor~\cite{wilson2023argoverse}. The nuScenes dataset covers diverse urban driving scenarios collected in Boston and Singapore. We follow the standard trajectory prediction protocol (6\,s future given 2\,s history) using the official split ($\approx$32k training / 9k validation samples).

The Argoverse~2 Sensor dataset provides large-scale multi-modal sensor data across six U.S.\ cities. Because it does not pre-define prediction targets, we extract samples by selecting valid agents based on object type, motion patterns, trajectory validity, and proximity to the ego vehicle, yielding 35k training and 7k validation samples under the same 2\,s\,/\,6\,s setting.

\subsubsection{Evaluation Metrics}
We report \textbf{minADE$_K$} (minimum Average Displacement Error over $K$ predicted modes), \textbf{minFDE$_K$} (minimum Final Displacement Error), and \textbf{Miss Rate (MR)} (fraction of predictions whose final displacement exceeds a threshold). All metrics are lower-better. For both datasets we use $K\!\in\!\{5,10\}$.

\subsection{Implementation Details}

All agent features are represented in the target-centric coordinate frame. The model uses a uniform hidden dimension $D\!=\!256$ across all modules. Pairwise geometric features (relative displacement, distance, relative velocity, approaching speed, longitudinal/lateral offsets) are projected to per-head attention logit biases via a two-layer MLP, enabling structured geometric reasoning without modifying the attention architecture. The scene-grounded relation module applies top-$k$ neighbor selection with distance-aware attenuation and residual refinement to stabilize training.

The model is implemented in PyTorch and trained with AdamW (weight decay $10^{-2}$) at an initial learning rate of $2\!\times\!10^{-4}$, linearly warmed up for 5 epochs followed by cosine annealing for 15 epochs (batch size 12, three NVIDIA A100 GPUs). The BEVFusion sensor encoder is kept frozen throughout trajectory-prediction training. For nuScenes, agent history is interpolated from 2\,Hz to 10\,Hz to match the Argoverse~2 Sensor sampling frequency.

\subsection{Comparison with State-of-the-Art Methods}

\subsubsection{nuScenes}

\begin{table}[t]
\centering
\caption{Performance on the nuScenes validation set. \textbf{MF}: map-free (no HD map at inference).}
\label{tab:nuscenes}
\setlength{\tabcolsep}{2pt}
\begin{tabular}{l|c|ccccc}
\hline
Method & MF & mADE$_5$$\downarrow$ & mADE$_{10}$$\downarrow$ & mFDE$_1$$\downarrow$ & mFDE$_{10}$$\downarrow$ & MR$\downarrow$ \\
\hline
Autobot~\cite{girgis2022autobot}        & $\times$ & 1.9566 & 1.1649 & 8.8171 & 2.3294 & 0.3229 \\
MTR~\cite{shi2022mtr}                   & $\times$ & 1.2926 & 1.0446 & 7.2689 & 2.2840 & 0.4240 \\
Wayformer~\cite{nayakanti2023wayformer} & $\times$ & 1.4034 & 0.9877 & 7.9707 & 2.2483 & 0.3868 \\
DeMo~\cite{zhang2024decoupling}         & $\times$ & 1.3137 & 1.0424 & 7.1679 & 2.1806 & 0.3399 \\
BEVTraj~\cite{kong2025bevtraj}          & $\checkmark$ & 1.4556 & 0.9438 & 8.4384 & 2.0527 & 0.3082 \\
\hline
\textbf{SIPTraj (Ours)} & $\checkmark$ & \textbf{1.1367} & \textbf{0.8766} & \textbf{6.6205} & \textbf{2.0079} & \textbf{0.2749} \\
\hline
\end{tabular}
\vspace{0.3em}
\end{table}

As shown in Table~\ref{tab:nuscenes}, SIPTraj achieves the best results across all five metrics among map-free methods. Compared with BEVTraj, the only other map-free BEV-based method in this comparison, SIPTraj reduces mADE$_5$ by 21.9\% (1.1367 vs.\ 1.4556), mADE$_{10}$ by 7.1\% (0.8766 vs.\ 0.9438), and Miss Rate by 10.8\% (0.2749 vs.\ 0.3082), demonstrating that hierarchical multi-stage scene grounding substantially outperforms single-step BEV fusion on both average accuracy and prediction consistency. The mFDE$_1$ improvement (6.6205 vs.\ 8.4384) further shows that the physics-guided decoder produces more accurate endpoint predictions for the best mode. Against map-based methods, SIPTraj surpasses Autobot and Wayformer on all metrics and outperforms DeMo on mADE$_5$ (1.1367 vs.\ 1.3137), mFDE$_{10}$ (2.0079 vs.\ 2.1806), and Miss Rate (0.2749 vs.\ 0.3399), demonstrating that multi-stage agent-scene coupling can effectively compensate for the absence of vectorized lane geometry.

\subsubsection{Argoverse 2 Sensor}

\begin{table}[t]
\centering
\caption{Performance on the Argoverse~2 Sensor validation set. \textbf{MF}: map-free.}
\label{tab:argo}
\setlength{\tabcolsep}{2pt}
\begin{tabular}{l|c|ccccc}
\hline
Method & MF & mADE$_5$$\downarrow$ & mADE$_{10}$$\downarrow$ & mFDE$_1$$\downarrow$ & mFDE$_{10}$$\downarrow$ & MR$\downarrow$ \\
\hline
Autobot~\cite{girgis2022autobot}        & $\times$ & 1.1417 & 0.6560 & 5.8954 & 1.3956 & 0.1762 \\
MTR~\cite{shi2022mtr}                   & $\times$ & 0.7880 & 0.6799 & 4.6435 & 1.7082 & 0.2821 \\
Wayformer~\cite{nayakanti2023wayformer} & $\times$ & 0.8191 & 0.5583 & 4.7608 & 1.3690 & 0.1837 \\
DeMo~\cite{zhang2024decoupling}         & $\times$ & 0.9000 & 0.6524 & 5.1404 & 1.3562 & 0.1776 \\
BEVTraj~\cite{kong2025bevtraj}          & $\checkmark$ & 0.9820 & 0.6249 & 5.2608 & 1.5832 & 0.1896 \\
\hline
\textbf{SIPTraj (Ours)} & $\checkmark$ & \textbf{0.7438} & \textbf{0.4635} & \textbf{4.8421} & \textbf{1.3422} & \textbf{0.1654} \\
\hline
\end{tabular}
\vspace{0.3em}
\end{table}

On Argoverse~2 Sensor (Table~\ref{tab:argo}), SIPTraj achieves the best results across all metrics among both map-free and map-based methods, demonstrating strong cross-dataset generalization of HASE and PGID. Against BEVTraj, SIPTraj reduces mADE$_5$ by 24.3\% (0.7438 vs.\ 0.9820), mADE$_{10}$ by 25.8\% (0.4635 vs.\ 0.6249), and Miss Rate by 12.8\% (0.1654 vs.\ 0.1896), with gains consistently larger than those on nuScenes. Against map-based methods, SIPTraj surpasses Autobot on all five metrics and achieves lower mADE$_5$ and MR than Wayformer and DeMo. Notably, SIPTraj achieves the lowest mFDE$_{10}$ (1.3422) among all methods, outperforming DeMo (1.3562) and Wayformer (1.3690), confirming that physics-aware feature conditioning improves long-horizon trajectory consistency even without lane-level map priors.

\subsection{Ablation Study}
\label{sec:ablation}

We conduct all ablations on the nuScenes validation set and report the same five metrics as the main comparison to enable direct cross-table comparison. The two groups isolate the contribution of HASE and PGID independently.

\subsubsection{Effect of Hierarchical Agent-Scene Encoding (HASE)}

We progressively enable each HASE stage to quantify the individual contribution of each coupling step. A1 applies Agent-to-Scene Aggregation (A2S) only, with the resulting scene memory average-pooled onto each agent token before decoding. A2 additionally enables Scene-to-Agent Grounding and Relation Refinement (S2A+Rel), which writes local scene evidence back into each agent representation and refines inter-agent interactions within the scene-grounded space. A3 is the full HASE, further incorporating Global Scene-Context Fusion (GSF), which aligns each agent token with the complete scene memory via cross-attention. All three configurations use the same decoder without PGID.

\begin{table}[t]
\centering
\caption{Ablation of HASE stages on nuScenes.}
\label{tab:ablation_hase}
\setlength{\tabcolsep}{1.5pt}
\renewcommand{\arraystretch}{0.95}
\begin{tabular}{l|ccc|ccccc}
\hline
Config & A2S & S2A & GSF & mADE$_5$$\downarrow$ & mADE$_{10}$$\downarrow$ & mFDE$_1$$\downarrow$ & mFDE$_{10}$$\downarrow$ & MR$\downarrow$ \\
\hline
A1         & $\checkmark$ & $\times$     & $\times$     & 1.3821 & 1.0112 & 7.9634 & 2.2841 & 0.3156 \\
A2         & $\checkmark$ & $\checkmark$ & $\times$     & 1.2754 & 0.9587 & 7.4218 & 2.1763 & 0.2998 \\
A3 (full)  & $\checkmark$ & $\checkmark$ & $\checkmark$ & 1.2031 & 0.9214 & 7.0852 & 2.1134 & 0.2883 \\
\hline
\multicolumn{9}{l}{\footnotesize A2S: Agent-to-Scene Aggr.; S2A: Scene-to-Agent+Rel; GSF: Global Fusion.}
\end{tabular}
\end{table}

Table~\ref{tab:ablation_hase} confirms that each HASE stage contributes incrementally across all five metrics. A2S alone (A1) conditions scene token extraction on agent motion context rather than reading the BEV map indiscriminately, achieving a meaningful improvement over a naive BEV-injection baseline across all metrics. Adding S2A+Rel (A2) provides the largest single-stage gain: mADE$_5$ drops from 1.3821 to 1.2754 and mFDE$_1$ from 7.9634 to 7.4218, as local scene evidence is written back into each agent token to spatially anchor it within the surrounding BEV context while relation refinement ensures proximity and directional cues are encoded before decoding. GSF (A3) further improves all metrics by allowing each agent to consult the complete scene memory rather than only its local neighborhood, with mADE$_5$ reaching 1.2031 and Miss Rate dropping to 0.2883, demonstrating that global structural context complements the local grounding established by S2A+Rel.

\subsubsection{Effect of Physics-Guided Iterative Decoder (PGID)}

Starting from the full HASE (A3), we ablate the two PGID components. B1 removes the Physics Query (PQ, i.e., PCAR) while retaining the Physics Consistency Losses (PL); B2 retains PQ while removing PL; B3 enables both and constitutes the complete SIPTraj model.

\begin{table}[t]
\centering
\caption{Ablation of PGID components on nuScenes. All rows use full HASE (A3). PQ\,=\,Physics Query via PCAR; PL\,=\,Physics Consistency Losses ($\mathcal{L}_a$+$\mathcal{L}_j$+$\mathcal{L}_c$).}
\label{tab:ablation_PGID}
\setlength{\tabcolsep}{2pt}
\begin{tabular}{l|cc|ccccc}
\hline
Config & PQ & PL & mADE$_5$$\downarrow$ & mADE$_{10}$$\downarrow$ & mFDE$_1$$\downarrow$ & mFDE$_{10}$$\downarrow$ & MR$\downarrow$ \\
\hline
B1 (w/o PQ)    & $\times$     & $\checkmark$ & 1.1842 & 0.9073 & 6.8934 & 2.0731 & 0.2812 \\
B2 (w/o PL)    & $\checkmark$ & $\times$     & 1.1756 & 0.8981 & 6.7810 & 2.0524 & 0.2793 \\
B3 (full PGID) & $\checkmark$ & $\checkmark$ & \textbf{1.1367} & \textbf{0.8766} & \textbf{6.6205} & \textbf{2.0079} & \textbf{0.2749} \\
\hline
\end{tabular}
\end{table}

\begin{figure}[t]
    \centering
    \includegraphics[width=\linewidth]{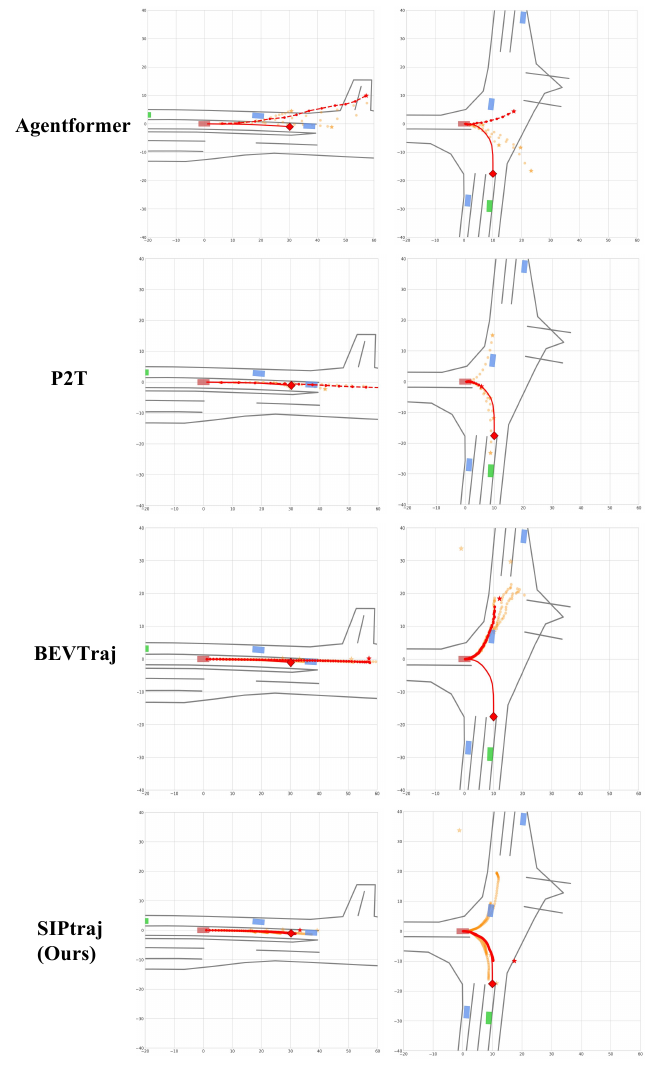}
    \caption{Qualitative comparison of predicted trajectories across straight-road and curved scenarios. Each column shows predictions from AgentFormer (top), P2T (middle), and SIPTraj (bottom). SIPTraj produces more accurate endpoint predictions and more road-consistent multimodal hypotheses, with distinct trajectory modes aligned to plausible driving intentions.}
    \label{fig:qual}
\end{figure}

Table~\ref{tab:ablation_PGID} confirms that PQ and PL are complementary and both necessary. B1 and B2 each improve over the HASE-only baseline (A3 in Table~\ref{tab:ablation_hase}), and their gains are comparable across all metrics. Without PQ (B1), consistency loss gradients are confined to the output coordinates and cannot propagate into the internal scene representation, limiting their effect to the behavior of prior output-level regularization~\cite{gao2024sif,zhong2025litransformer}. Without PL (B2), the PCAR cross-attention channel is structurally present but receives no physics-grounded training signal, leaving the physics query without meaningful supervision. Only B3 couples both: the consistency losses backpropagate through the PCAR cross-attention into the physics MLP, causing the internal physics encoding to co-adapt with the coordinate refinement objective. This co-adaptation is most visible in mFDE$_1$ (6.6205 vs.\ 6.8934 for B1 and 6.7810 for B2), where accurate endpoint prediction requires both a dedicated internal pathway and output-level physical supervision. The consistent improvement of B3 over B1 and B2 across all five metrics supports the design principle that feature-level modulation and output-level supervision are inseparable in PGID.

\subsection{Qualitative Analysis}
\label{sec:qualitative}

As shown in Fig.~\ref{fig:qual}, SIPTraj produces more compact mode distributions with probability mass concentrated near the highest-likelihood hypothesis and endpoint predictions consistently closer to the ground truth than AgentFormer and P2T in straight-road cases. In curved scenarios, AgentFormer and P2T generate modes that deviate from road curvature at longer horizons, while SIPTraj's hierarchical scene grounding keeps each mode spatially anchored to the local BEV context. The physics-guided decoder further suppresses kinematically implausible hypotheses, yielding a cleaner multimodal distribution with distinct, feasible maneuvers per mode.

\section{CONCLUSIONS}

We presented SIPTraj, a map-free trajectory prediction framework integrating hierarchical agent-scene encoding and physics-aware feature conditioning. The Hierarchical Agent-Scene Encoder couples agent-conditioned BEV scene encoding, local scene grounding, relation-aware interaction refinement, and late task refinement into a unified pipeline. The Physics-Guided Iterative Decoder (PGID) bridges physical 
supervision from the output level to the internal scene representation by conditioning prediction features on each agent's instantaneous kinematic state. Physical constraint losses on acceleration, jerk, and curvature further enforce trajectory feasibility. Experiments on nuScenes and Argoverse 2 Sensor demonstrate that SIPTraj achieves competitive map-free trajectory prediction performance, validating the effectiveness of jointly modeling scene context, structured interaction, and physical feasibility.

\addtolength{\textheight}{-12cm}   % This command serves to balance the column lengths

%%%%%%%%%%%%%%%%%%%%%%%%%%%%%%%%%%%%%%%%%%%%%%%%%%%%%%%%%%%%%%%%%%%%%%%%%%%%%%%%

%%%%%%%%%%%%%%%%%%%%%%%%%%%%%%%%%%%%%%%%%%%%%%%%%%%%%%%%%%%%%%%%%%%%%%%%%%%%%%%%

%%%%%%%%%%%%%%%%%%%%%%%%%%%%%%%%%%%%%%%%%%%%%%%%%%%%%%%%%%%%%%%%%%%%%%%%%%%%%%%%

%%%%%%%%%%%%%%%%%%%%%%%%%%%%%%%%%%%%%%%%%%%%%%%%%%%%%%%%%%%%%%%%%%%%%%%%%%%%%%%%

\bibliographystyle{ieeetr}
\bibliography{root.bib}

\end{document}